\pdfoutput=1

\documentclass[11pt]{article}

\usepackage[final]{acl}

\usepackage{times}
\usepackage{latexsym}

\usepackage[T1]{fontenc}

\usepackage[utf8]{inputenc}

\usepackage{microtype}

\usepackage{inconsolata}

\usepackage{graphicx}

\usepackage{CJK}

\usepackage{algorithm2e}
\usepackage{amssymb}
\usepackage{marvosym}
\usepackage{hyperref}

\def \yuan #1{{\color{black}#1}}

\title{Can Large Language Models Grasp Legal Theories? \\Enhance Legal Reasoning with Insights from Multi-Agent Collaboration}

\author{
 \textbf{Weikang Yuan\textsuperscript{1,2}},
  \textbf{Junjie Cao\textsuperscript{2}},
 \textbf{Zhuoren Jiang\textsuperscript{1 }\thanks{Corresponding author.}},
 \textbf{Yangyang Kang\textsuperscript{1}},
 \textbf{Jun Lin\textsuperscript{2}},
  \\
 \textbf{Kaisong Song\textsuperscript{3,2}},
 \textbf{Tianqianjin Lin\textsuperscript{1,2}},
 \textbf{Pengwei Yan\textsuperscript{1,2}},
 \textbf{Changlong Sun\textsuperscript{2}},
 \textbf{Xiaozhong Liu\textsuperscript{4}}
\\
 \normalsize{\textsuperscript{1}Zhejiang University, \textsuperscript{2}Tongyi Lab, Alibaba Group}
 \\
  \normalsize{\textsuperscript{3}Northeastern University, \textsuperscript{4}Worcester Polytechnic Institute}
\\
\texttt{
\small{
\{yuanwk, jiangzhuoren, yangyangkang, lintqj, yanpw\}@zju.edu.cn, \{junjie.junjiecao, linjun.lj,}
}
\\
\texttt{
\small{kaisong.sks\}@alibaba-inc.com, }
   changlong.scl@taobao.com, xliu14@wpi.edu
}
}

\begin{document}
\maketitle
\begin{abstract}
Large Language Models (LLMs) could struggle to fully understand legal theories and perform complex legal reasoning tasks. In this study, we introduce a challenging task (confusing charge prediction) to better evaluate LLMs' understanding of legal theories and reasoning capabilities. We also propose a novel framework: \underline{M}ulti-\underline{A}gent framework for improving complex \underline{L}egal \underline{R}easoning capability (MALR). MALR employs non-parametric learning, encouraging LLMs to automatically decompose complex legal tasks and mimic human learning process to extract insights from legal rules, helping LLMs better understand legal theories and enhance their legal reasoning abilities. Extensive experiments on multiple real-world datasets demonstrate that the proposed framework effectively addresses complex reasoning issues in practical scenarios, paving the way for more reliable applications in the legal domain.
\end{abstract}

\section{Introduction}
Large Language Models (LLMs) have shown remarkable generalization ability across diverse range of tasks and applications~\citep{chowdhery2023palm,touvron2023llama,achiam2023gpt}. But, current benchmarks may not adequately reflect the reasoning capabilities of LLMs~\citep{valmeekam2024planbench} and do not accurately reflect real-world situations~\citep{huang-2023-LLM-Reasonig-Survey}. The validation of LLMs in more realistic and meaningful applications, such as legal reasoning, still requires extensive exploration.

\begin{figure*}[t]
  \includegraphics[width=0.85\linewidth]{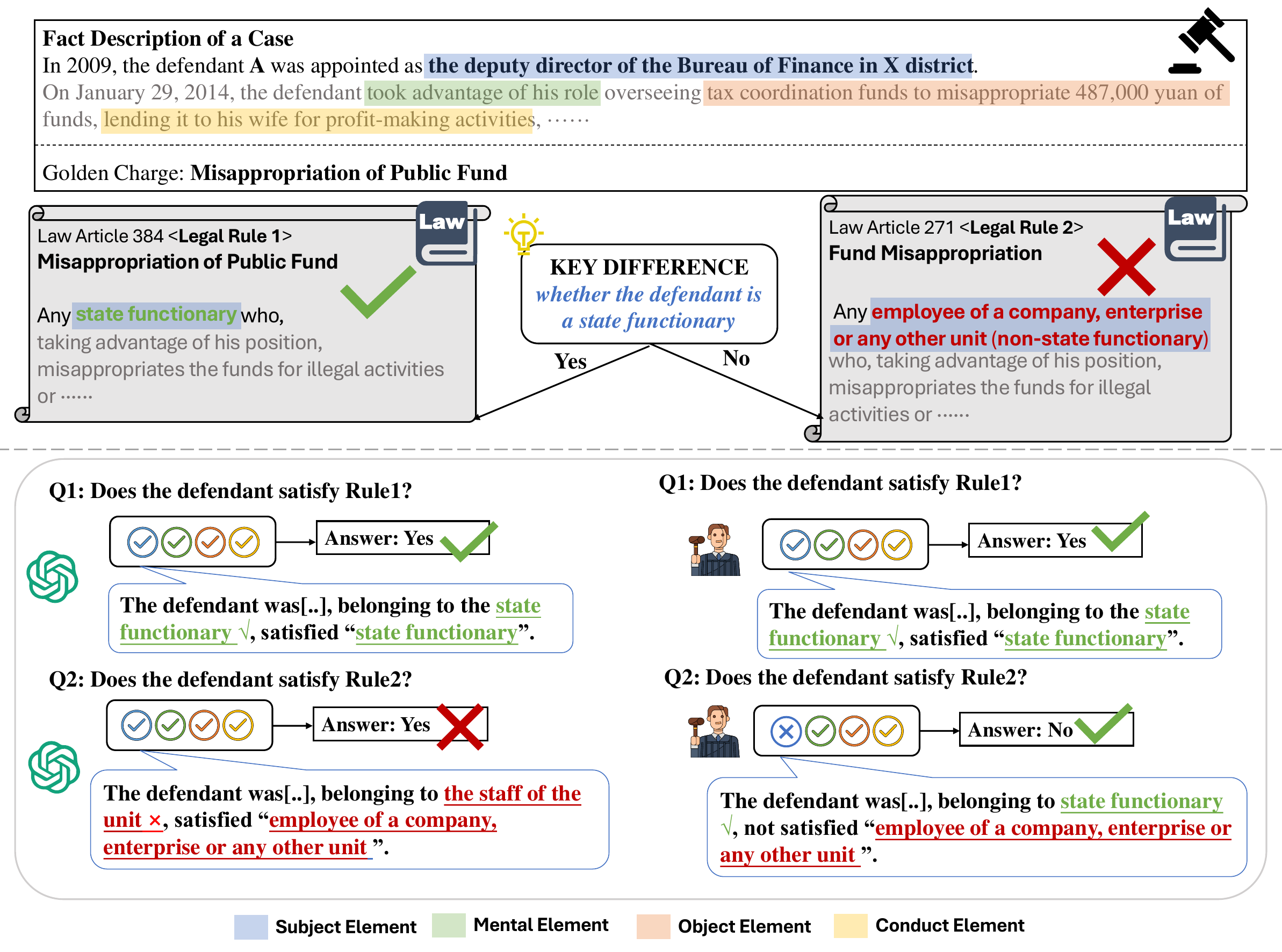} 
  \centering
  \caption {An example to demonstrate how a judge and an LLM apply legal rules to conclude whether a case satisfies a specific charge. This example outlines two confusing charges under Chinese criminal law: the \textbf{Crime of Fund Misappropriation} and the \textbf{Crime of Misappropriation of public fund}. The most significant difference between the two charges is \textcolor{blue}{whether the defendant is a state functionary}. In the case description, the defendant is \textit{``the Deputy Director of the Bureau of Finance in X district''}, a position that qualifies as a state functionary. Therefore, the judge can easily infer that the case falls under the Crime of Misappropriation of public funds, rather than Fund Misappropriation. However, the LLM fails to predict the confusing charge.
}
    \label{fig:toyexample}
\end{figure*}

In the legal domain, the core competency of legal professionals is to apply legal rules to facts and draw conclusions, as described by the IRAC (Issue, Rule, Application, Conclusion) framework. As shown in Figure \ref{fig:toyexample}, a legal professional can determine whether a case fact conforms to specific criminal charges based on legal rules. They critically assess a case against potential charges, focusing on the key points of relevant legal rules, to accurately identify the appropriate charge and distinguish inapplicable charges. Legal rules, which manifest legal theories, determine the legal consequences of factual situations~\citep{maccormick2005rhetoric}. Therefore, properly applying legal rules reflects the grasp of legal theories.

\begin{figure}[t]
  \includegraphics[width=\columnwidth]{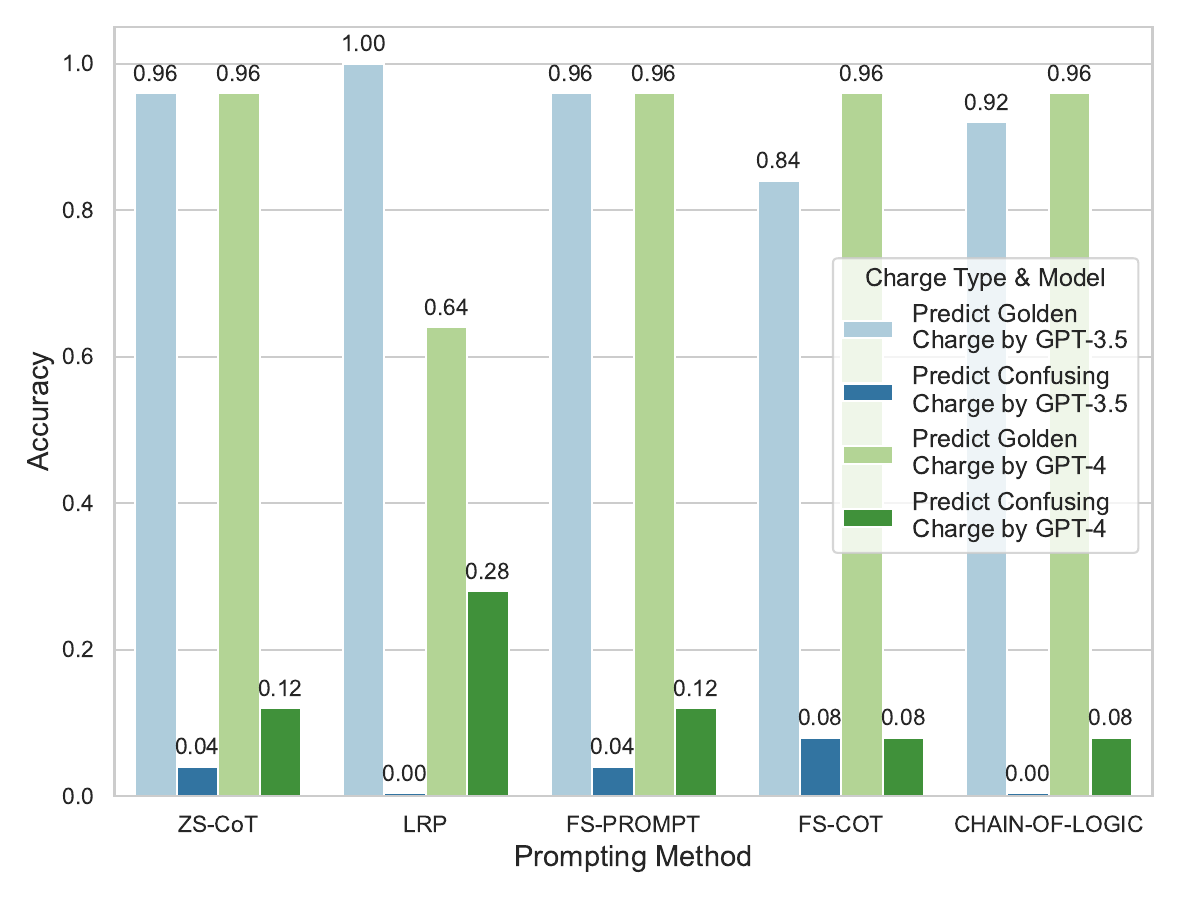}
  \caption{The performance of LLMs on predicting the golden (\textbf{Misappropriation of Public Fund}) or confusing charge (\textbf{Fund Misappropriation}) for the cases from CAIL-2018 datasets. The horizontal axis represents 5 advanced promt methods to solve legal reasoning problems (detailed information is described in Section \ref{section:Experiments}). In each method, GPT-3.5 and GPT-4 both exhibit a significant performance gap.}
  \label{fig:LLM LIMITATION}
\end{figure}

However, powerful LLMs may struggle to fully understand legal theories and perform basic legal reasoning tasks. Existing study~\citep{dahl2024largelegalfiction} has found that when LLMs are given criminal facts and legal rules, then asked whether cases constitute a certain charge, they tend to answer ``yes'', regardless of whether the charge is correct (golden charge) or a closely related one  (confusing charge). 
Our empirical experiments also confirmed this issue. We sampled real-world criminal cases involving the charge of \textbf{Misappropriation of Public Fund}, inputting the criminal facts and legal rules into LLMs, and asked whether the case constituted the golden charge. Meanwhile, we created a control group where we input the same criminal facts and related legal rules, asking whether the case constituted a confusing charge  (\textbf{Fund Misappropriation}). These two charges are very similar, with the key difference being \textit{whether the defendant's subject position is that of a state functionary}. As shown in Figure \ref{fig:LLM LIMITATION}, when performing legal reasoning, regardless of the prompt method or the version of GPT used, LLMs exhibit significant declines in performance when predicting confusing charges.

Generally, LLMs could face following challenges in legal reasoning:
\textbf{Inconsistent reasoning.} Legal reasoning involves multi-step, compositional logic processes~\citep{servantez2024chain}.
LLMs can be easily distracted by the interaction when generating reasoning steps~\citep{LLMdistracted} and may not be trustworthy by the tendency to give affirmative answers~\citep{dahl2024largelegalfiction}.
\textbf{Missing key details.} Legal rules and criminal facts are often described in complex natural language, making it challenging for LLMs to fully understand and reason based on them. Consequently, they often overlook key information in the rules.
\textbf{Lacking domain knowledge.} LLMs may hallucinate erroneous legal knowledge~\citep{dahl2024largelegalfiction} or encounter gaps in common-sense knowledge~\citep{huang2023survey}. Their overconfidence can obscure these shortcomings, making them difficult to identify~\citep{ni2024llms}.

To better evaluate LLMs' understanding of legal theories and their reasoning capabilities, we introduce and construct a challenging task: confusing charge prediction (The detailed task definition is provided in Section \ref{sec:task}). We also propose a novel framework: \underline{M}ulti-\underline{A}gent framework for improving complex \underline{L}egal \underline{R}easoning capability (MALR). First, an auto-planner breaks down complex legal rules into sub-tasks, allocating them to expert agents, reducing inconsistent reasoning in LLMs. Second, a non-parametric learning framework is proposed to draw adaptive rule-insights from trials and errors. To address the problem that LLMs may overlook crucial information in legal rules, we design a module that mimics human learning by gaining experience through reasoning trajectories and knowledge feedback, then learning insights through self-reflection. These insights supplement the rules, encouraging LLMs to focus on key factors from legal knowledge and fully understand the rules, while also guiding them to automatically seek help when they feel uncertain. These designs effectively improve LLMs' reasoning and critical-thinking skills. 

Our contributions are threefold:

$\bullet$  We propose a multi-agent framework based on non-parametric learning, which encourages LLMs to automatically decompose complex legal tasks and extract insights from legal rules. Our framework assists LLMs in gaining a deeper understanding of legal rules and enhances their legal reasoning capabilities.

$\bullet$  We introduce a challenging task, confusing charge prediction, to better evaluate LLMs' understanding of legal theories and their reasoning capabilities.

$\bullet$  Extensive experiments are conducted on the multiple real-world datasets, demonstrating that the proposed framework can effectively addresses complex reasoning issues in real-world scenarios. Our work paves the way for more trustworthy application in legal domain\footnote{Our source code can be found at \url{https://github.com/yuanwk99/MALR}}.

\section{Related Work}
\subsection{Legal AI and LLMs }
Legal AI aims to improve legal tasks through AI techniques, particularly showing significant potential in alleviating  the issue of ``too many cases but too fewer legal experts'' in the legal field~\citep{katz2023natural,dahl2024largelegalfiction}.
One of the main challenges in legal domain is the training dataset can be considerably expensive and sparse~\citep{legal-problem}, primarily comes in text, such as statutes, law articles and criminal cases.
Under these circumstances, LLMs shows promising prospects in legal scenarios due to their powerful generalization capabilities in understanding and generating text.
These applications include areas such as legal summarization~\citep{deroy2023ready},  legal document retrieval~\citep{sun-etal-2024-logic-rules}, legal question answering~\citep{louis2024interpretable} and legal judgment prediction~\citep{yu2022legalprompting,CJO,servantez2024chain}.

\subsection{Legal Reasoning and LLMs}
Reasoning based on judicial rules and case fact descriptions is a fundamental ability of legal professionals, reflecting their understanding and application of legal theories~\citep{servantez2024chain}. Previous studies on Legal Judgment Prediction (LJP) have primarily focused on automatically predicting case charges mainly from fact descriptions~\citep{zhong2018legal,chalkidis-etal-2019-neural,liu2022augmenting}. Additionally, similar precedents can be retrieved as supplementary guides to improve performance~\citep{CJO}. However, this approach can lead to inaccurate judgments due to overlooked potential differences in case details. To address the subtle differences between case details and legal rules, knowledge graphs have been introduced to solve confusing charges problems~\citep{yue2021neurjudge,li-etal-2024-graph-word}. Despite these efforts, utilizing Four Elements Theory and innocent datasets, \citet{CAIL-I} found that charge prediction models do not take legal theories into consideration. Instead, models learn certain shortcuts for legal reasoning. Furthermore, Chain-of-Logic~\citep{servantez2024chain} directly incorporates legal rules into prompts to elicit rule-based reasoning, achieving good performance on legal reasoning tasks involving three distinct rules from the LegalBench benchmark~\citep{guha2024legalbench}. SimuCourt proposes a multi-agent framework to simulate the decision-making process of a judge~\citep{he2024simucourt}.

Unlike existing works, we aim to evaluate and enhance the capacity of LLMs to reason based on legal theories, rather than treating legal rules as supplementary information.

\section{Preliminary}\label{sec:task}

We propose \textbf{Confusing Charge Prediction Task} to evaluate the LLMs' ability to identify correct legal charges based on fact descriptions and legal rules.
\yuan{This task highlights subtle distinctions in legal rules. Only LLMs that capture these nuances can demonstrate their understanding of legal theory.}

\textbf{Fact Description}: a concise description of a legal case, represented as a word sequence $\textbf{f} = \{w_1,w_2,...,w_l\}$. \textbf{Legal Rule}: the definition of a specific criminal charge from law articles, also a word sequence $\textbf{r}_c = \{w_1,w_2,...,w_n\}$, where $c$  is the criminal charge. \textbf{Golden Charge}: The true crime label of a case. \textbf{Confusing Charge}: A charge similar to the golden charge but differing in one element~\citep{CAIL-I}.

To ensure LLMs' trustworthiness in applying legal rules, we require them to confirm the golden charge as True and reject the confusing charge as False.
The task can be formalized as:
 $$y = \Gamma(f, r_{gc}) \land \neg \Gamma(f, r_{cc})$$
where $gc$ refers to the golden charge, $cc$ refers to the confusing charge, and $\Gamma$ is the charge prediction model. $y$ is True only if the fact description $f$ satisfies the rule of golden charge $r_{gc}$ and does not match the rule of confusing charge $r_{cc}$.

LLMs should correctly identify the golden charge and explain why the fact description doesn't match the confusing charge, demonstrating understanding of legal theories.

\section{The Proposed Framework}
\begin{figure*}[t]
  \includegraphics[width=0.88\linewidth]{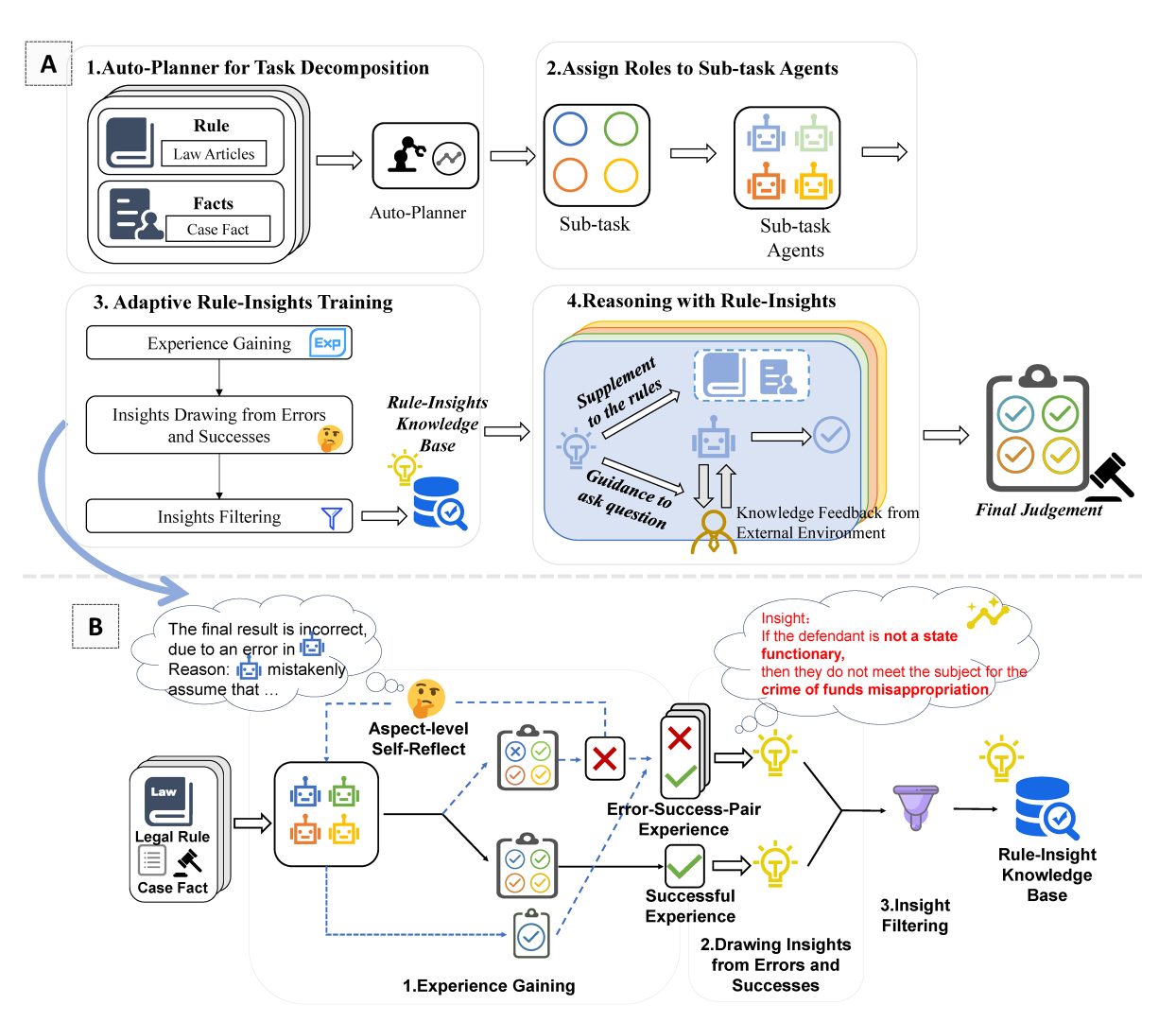} 
  \centering
  \caption {Our research framework in (A) and Adaptive Rule-Insights training process in (B). }
  \label{fig-framework}
\end{figure*}

Figure \ref{fig-framework} shows an overview of our proposed framework, which consists of four core components: Auto-Planner for Task Decomposition, Role Assignment for Sub-task Agent, Adaptive Rule-Insights Training, and Reasoning with Rule-Insights.

\subsection{Auto-Planner}
A single LLM may exhibit inconsistencies when directly generating the whole reasoning process~\citep{wang2024can}. Therefore, we designed an automatic planning module to decompose the task. Given a question $q$, a case fact description $\textbf{f}$, and the corresponding legal rule $\textbf{r}_c$ about a criminal charge $c$, we guide an LLM as \textit{auto-planner} to decompose the question into a sequence of sub-tasks based on the input of the fact and the rule:
\begin{equation}
  \label{autoplanner}
  [st_1,...,st_k] =  LLM(q,\textbf{r}_c,\textbf{f},p
_{auto})
\end{equation}
where $p_{auto}$ is the guideline prompt for LLMs to generate the sub-task set for the question $q$, and the $st$ stands for the specific sub-task action and the $k$ is the length of the sub-task set.

Given the resource-intensive nature of generating sub-tasks for every criminal case, we design a more effective strategy. We first sample a smaller scale dataset for auto-planner training, and prompt the LLM to generate the sub-task set for each sample. Subsequently, an LLM is used to identify semantically duplicate sub-task and compute the probability distribution for each sub-task. Based on this process, important sub-tasks with probability exceeding the threshold $\zeta$ are used to constitute the final sub-task set.

\subsection{Assigning Roles to Sub-task Agent}
Assigning roles can help the LLMs better perform complex task reasoning~\citep{wang2023rolellm}. Therefore, based on the sub-task set $[st_1,...,st_k]$, we employ k LLM-based agents to tackle each sub-task. Formally, we use the content of the sub-tasks to generate the appropriate prompts $p_{st}$ and generate k agents to tackle each sub-task. Each agents will break down specific aspects of legal rule, check whether the fact description $f$ conforms the corresponding sub-rule $r_{c_{st}}$ and generate the answer $A_{st}$.
This process can be formulated as:
$$A_{st} = M_{st}(r_{c_{st}},f,p_{st})$$
\yuan{After obtaining the answer for each sub-task, a logic expression~\citep{servantez2024chain} based on the principle of "presumption of innocence"~\citep{CAIL-I} will generate the final answer.
}

\subsection{Adaptive Rule-Insights Training}
As aforementioned, LLMs can be easily distracted by the irrelevant context~\citep{LLMdistracted} and tend to overlook the key information and important details within rules.
Therefore, we aim to enable LLMs to automatically extract the most critical information for legal judgement directly from the rules.
Previous research demonstrated that LLMs can autonomously learn from their own experiences~\citep{shinn2024reflexion,zhao2024expel}.
Inspired by the Kolb's Experiential Learning Model~\citep{kolb2014experientiallearningmodel}, we design the insights training module, as shown in Figure \ref{fig-framework} (B), which consists of three core processes: experience gaining, insights drawing from errors and successes, and insights filtering. This module mimics the human learning process and facilitates the LLMs to automatically learn rules, discovering and summarizing the most critical information in the rules.

\noindent \textbf{Experience Gaining}. A small training dataset with $N$ charges is constructed, with each charge containing case samples and corresponding confusing charges. Based on the fact descriptions of a case,  sub-task agents $M_{st}$ will respectively generate sub-answers for both golden charge and confusing charge. These sub-answers will be synthesized into a final determination of whether the case satisfies the legal rule for the golden charge or confusing charge, and the ground truth is used as feedback.
Successful trials are saved as successful experience, while failed trials trigger the Aspect-level Self-Reflector to identify incorrect sub-task agents $M_{st}^{e}$ and generate reasons $rs_e$ for the errors. In the next trial, the error reasons are used to improve sub-task agents' predictions. Such approach of learning from trials and errors can be effective, as demonstrated in~\citep{shinn2024reflexion}. This iterative process continues for a maximum of $L$ rounds, and corrected trajectories are retained as error-success-pair experience. The algorithm procession is detailed in Alg \ref{alg:insight}.


\renewcommand{\algorithmcfname}{Alg}

\RestyleAlgo{ruled}
\SetKwComment{Comment}{/* }{ */}

\begin{algorithm}[hbt!]
\caption{Experience Gaining}\label{alg:insight}
\textbf{Initialze:}Self-Reflector, Sub-task Agent, Evaluator: $M_{reflect},M_{st},M_{e}$\\
Number of charges $N$\\
Successful Experience $E_{success}$\\
Error-Success-Pair Experience $E_{esp}$\\
trajectory $\tau$, Fact Description $f$\\
Golden Charge $gc$, Confusing Charge $cc$ 
\While{charge $n \leq N$}{
Set $r_{gc},r_{cc} \gets gc_n,cc_n$\\
Generate $\tau_{l,gc} = [A_{1,gc},..A_{k,gc}]$ using $M_{st},r_{gc},f$\\
Generate $\tau_{l,cc} = [A_{1,cc},..A_{k,cc}]$ using $M_{st},r_{cc},f$\\
\While{trial $l \leq L$}{
Evaluate ($\tau_{l,gc},\tau_{l,cc}$) using $M_e$\\
    \eIf{success}{
        \eIf{$l=1$}{
            Append $\tau_{l,gc},\tau_{l,cc}$ to $E_{success}$\\
            break
        }
        {
         Append $\tau_{l,gc},\tau_{l,cc}$ to $E_{esp}$\\
         break
        }
    }
    {
    Identify error $M^e_{st}$ and Generate ${rs}_e$ using  $M_{reflect}$\\
    $e \in \{gc,cc\}$\\
    Generate $A_{k,e}$ using $M^e_{st}, r_e, f, rs_e$\\
    Updating $\tau_{l+1,e}$ using $ A_{k,e}$
    }
}
}
\end{algorithm}

\noindent \textbf{Insights Drawing from Errors and Successes}. We gain insights into rules by analyzing experience collections using different methods. For error-success pairs, we use a contrast-based approach, comparing incorrect and correct attempts. This enables the $M_{insight}$ to identify the most critical task-level judgments where errors occur, outputting insights in an if-then format. Successful experiences reveal best practices~\citep{zhao2024expel}, so we provide the entire successful reasoning process to $M_{insight}$ to generate corresponding insights.

\noindent \textbf{Insights Filtering}. To address the potential for generating duplicate or incorrect insights when interpreting rules from the aforementioned process, we employ an LLM as an automatic checker, $M_{filter}$, to identify and remove redundant insights and filter out invalid ones. Ultimately, the retained insights are stored in the rule-insight knowledge base as memory. The pseudo-code for insight drawing and filtering is presented in Algorithm \ref{alg:insight-2}.

\begin{algorithm}[hbt!]
\caption{Insight Drawing and Filtering}\label{alg:insight-2}
\textbf{Initialze:}Insight-Drawer, Insight-Filter: $M_{insight},M_{filter}$\\
Successful Experience $E_{success}$\\
Error-Success-Pair Experience $E_{esp}$\\
Number of charges $N$\\
Number of sub-task $k$\\
Rule-Insight Knowledge Base $I$\\
\While{charge $n \leq N$}{
Construct error-success pair of sub-task trial from $E_{esp}$:
$\mathbf{P} = \{(A_{st_k}^{error},A_{st_k}^{success}),...\}$\\
\For{each $p$ in $\mathcal{P}$}{
Drawing insight $i$ using $M_{insight}(p)$\\
Update $i$ to $I[charge][st_k]$
}
\For{each $exp$ in  $E_{esp}$}
{
Drawing insight $i$ using $M_{insight}(exp)$\\
Update $i$ to $I[charge][st_k]$
}
Filter $I[charge]$ using $M_{filter}$
}

\end{algorithm}

\subsection{Reasoning through Insights }
The generated insights serve two purposes:
(1) they supplement the rules as additional notes, and
(2) they guide LLMs to inquire about uncertainties when facing knowledge gaps in specific sub-tasks.

For implementation, we retrieve relevant insights ${in}_{st}$ from the knowledge base $I$ for each question to improve reasoning. If the rule does not exist, the most similar rule from the knowledge base is selected based on semantic similarity, and a few-shot method is used to generate new insights.
To address potential knowledge gaps in LLMs, our insights serve as guidance to ask factuality questions.
Based on the insights, we identify sub-tasks requiring fact-checking and use a few-shot method to prompt LLMs to ask key questions like ``Is a <job position> a <state functionary>?''
The generated question is then presented to a knowledgeable expert (a legal professional, a domain-specific LLM, or a search engine) to obtain knowledge feedback ${kg}_{st}$.
Finally, we incorporate related insights ${in}_{st}$ and knowledge feedback ${kg}_{st}$ into our ultimate reasoning process.
As shown in Figure \ref{fig-framework}(A)(4), the improved reasoning process for each sub-task agent can be represented as:
\begin{equation}
      \label{finalreason}
A_{st} = M_{st}(r_{st}, f,in_{st},kg_{st},p_{st})
\end{equation}
All prompt templates for our MALR agents are provided in Appendix \ref{sec:appendix MALR Template}.

\section{Experiment}
\label{section:Experiments}
\subsection{Experimental Setting}
\textbf{Dataset}. We evaluate LLMs' legal reasoning capability on three datasets:
CAIL2018~\citep{xiao2018cail2018}, CJO~\citep{CJO}, and CAIL-I~\citep{CAIL-I}.
CAIL2018 and CJO consist of real-world cases with fact descriptions and golden charges.
We match corresponding confusing charges based on the golden charges and randomly sample 400 cases from CAIL2018 and 100 from CJO for evaluation.
CAIL-I's testset contains 462 innocent cases without crimes and the most similar charges to each non-criminal fact.
Further dataset information is available in Appendix \ref{sec:appendix dataset}.

The pairs of confusing charges are carefully selected by a group of legal experts, including: (1) Misappropriation of Public Fund (MP) v.s. Fund Misappropriation (FM); 
(2) Bribery (BY) v.s. Bribery of Non-State Officials (BN); 
(3) Kidnapping (KD) v.s. Illegal Detention (ID); 
(4) Fraudulently Obtaining Loans (FL) v.s. Loan Fraud (LF); 
(5) Fund Misappropriation (FM) v.s. Official Embezzlement (OE); 
(6) Fraud (FD) v.s. Loan Fraud (LF); 
(7) Fraud (FD) v.s. Cheating and Bluffing (CB); 
(8) Forging, Altering, Trading Official Documents, Certificates and Seals of State Organs (FO) v.s. Forging the Seals of Companies, Enterprise, Institution (FS). Key differences between each pair are provided in Appendix \ref{sec:appendix dataset}.

\noindent \textbf{Implementation}. We employ the publicly available GPT-3.5-Turbo-0125 and GPT-4-0125-preview models, with the temperature set to 0 for all text generation tasks. For auto-planner and insights training, we construct a small training set from CAIL-2018 training set. Specifically, we sample two cases for each of the 16 charges (totally 32 training samples). The threshold $\zeta$ for the sub-task auto-planner is set to 0.8. Sentence-BERT~\citep{thakur-2020-AugSBERT} and cosine similarity are used to compute semantic distances between rules and unseen legal rules, facilitating rule-insight inference testing in CJO and CAIL-I. During the insights training period, we limit the maximum number of trial attempts $L$ to 2.
For providing knowledge feedback, we employ Farui-200B\footnote{A legal-domain fine-tuned LLM based on Qwen\citep{qwen}, \url{https://tongyi.aliyun.com/farui}.}, which can be replaced by other legal models or even legal experts in real-world scenarios.
Additionally, we construct a legal rule knowledge base that includes Chinese Criminal Law Articles and all charge definitions. All legal rules are retrieved from this knowledge base based on the charge name. The inference time and cost can be seen in Appendix \ref{sec:appendix dataset}.

\subsection{Baselines}
\textbf{Zero-shot Setting}: (1) ZS-CoT~\citep{ZS-COT} uses ``\textit{Let's think step by step}'' to encourages LLMs to generate intermediate steps and improve reasoning. (2) Legal Reasoning Prompting (LRP)~\citep{yu2022legalprompting} is a zero-shot legal prompting method that teaches LLMs to reason like a lawyer, following the ``Approach, Issue, rule, application and conclusion'' framework.

\noindent \textbf{Few-shot Setting}: (1) Few-Shot prompting~\citep{brown2020language-few-shot} is the standard prompting method includes only the sample and answer. We use a two-shot setting with one positive and one negative examples. (2) Few-Shot CoT~\citep{wei2022chain} uses a few chain-of-thought demonstrations as exemplars to improve the ability of LLMs to perform complex reasoning. Again, we employ a two-shot setting with one positive and one negative examples. (3) Chain-of-Logic~\citep{servantez2024chain} elicits rule-based reasoning by decomposing the rule into elements, answering each rule element separately, and finally using a logical expression to obtain the final answer. This approach is meticulously designed for legal reasoning tasks.

All prompt template can be seen in Appendix \ref{sec:appendix Baseline Template}.

\begin{table*}[h]
  \centering
   \scalebox{0.8}{
  \begin{tabular}{c|cc|cc|cc} 
     \hline
    Methods& \multicolumn{2}{c}{CAIL2018}
& \multicolumn{2}{c}{CJO} 
&  \multicolumn{2}{c}{CAIL-I} 
\\ 
 \cline{2-7}
 & GPT-3.5& 
GPT-4& GPT-3.5
& GPT-4
& 
GPT-3.5&GPT-4\\ 
 \hline
    
    ZS-CoT~\citep{ZS-COT}& 12.5&                           35.8
&3.0
& 29.0
& 20.9 
&36.0\\ 
    LRP~\citep{yu2022legalprompting}& 9.8&                           37.8
&1.0
& 37.0
& 22.3&49.6\\ 
    \hline
    FS-Prompt~\citep{brown2020language-few-shot}& 18.0&                           41.0
&3.0
& 43.0
& 28.1 
&46.8\\ 
    FS-CoT~\citep{wei2022chain}& 12.0&                            
34.0& 
12.0& 
18.0& 
12.2&
32.4\\ 
    Chain-of-Logic~\citep{servantez2024chain}&                         6.5&  36.0
& 5.0& 25.0& 10.1
&
29.5\\ 
   \hline 
 MALR w/o insight& 32.3&43.8
&22.0
& 44.0
& 45.3&53.2\\ 
     MALR w/o ask& \underline{37.3} &\underline{53.3}
&\underline{31.0}
&\underline{ 53.0}
& \underline{51.1} & \underline{55.4}\\ 
  MALR (our)& \textbf{40.8} &\textbf{56.8}& \textbf{39.0}& \textbf{55.0}& \textbf{56.1}&\textbf{57.6}\\ 
 \hline
  \end{tabular}
  }
  \caption{\label{Main-result}
    Main results on three legal datasets, the best is \textbf{bolded} and the second is \underline{underlinded}. The metric is accuracy. w/o insight refers to only decompose to sub-tasks, w/o ask refers to do not get any external knowledge feedback.
  }

\end{table*}

\subsection{Experiment Results}
\noindent \textbf{Main Results:} From Table \ref{Main-result}, we observe the following findings.
(1) LLMs fail to distinguish confusing charges using simple but effective prompt methods such as CoT, and legal-specific prompting approaches also fail to predict accurately. By examining the actual prediction results, we found that LLMs using these methods tend to respond with ``yes.'' 
(2) ``MALR w/o insight'', which only decomposes the task into sub-tasks, outperforms all the baselines. This result indicates that decomposing the task into sub-tasks may mitigate LLMs' biased tendencies. Notably, without any human intervention, our auto-planning strategy can decompose legal rules into four aspects: Subject (Sub), Mental (Men), Object (Obj) and Conduct (Con). This aligns with the Four Elements Theory~\citep{CAIL-I}, which is widely recognized in the legal domain.
(3) ``MALR w/o ask'' does not utilize external knowledge feedback but still achieves the second-best results, indicating that the learned insights did significantly enhance the LLM's understanding of legal rules.
(4) The complete MALR achieves the best performance on all datasets, demonstrating the effectiveness of proposed framework and the necessity of its core components. 
MALR achieves the best performance on nearly all confusing charge pairs (refer to Appendix  \ref{sec:appendix detail}).
(5) Regarding the base models, GPT-3.5 benefits more from our proposed MALR compared to GPT-4, achieving a more significant improvement over the baseline methods. This suggests that our framework has a stronger enhancing effect on LLMs with weaker foundational capabilities.

\begin{table}[h]
  \centering
   \scalebox{0.8}{
  \begin{tabular}{c|cc|cc}
    \hline
    Datasets&  \multicolumn{2}{c}{CAIL2018}&\multicolumn{2}{c}{CJO} \\
 Methods& GPT-3.5& \textbf{GPT-4}  & GPT-3.5&\textbf{GPT-4}  \\
    \hline
    w/o insights&  32.3&43.8  & 22.0&44.0\\
    w/o $E_{success}$&  38.8 &50.0  & 29.0&48.0\\
    w/o $E_{esp}$&  46.0& 48.8 & 33.0&48.0\\
    w/o $M_{filtering}$&  38.0&54.0  & 31.0&53.0\\
 directly generate& 32.0& 43.3& 35.0&38.0\\
 complete MALR& 40.8& 56.8& 39.0&55.0\\ \hline
  \end{tabular}
}
  \caption{Ablation test for adaptive rule-insights training module.}
  \label{tab:ablation}
\end{table}

\begin{figure*}[h]
  \includegraphics[width=0.8\linewidth]{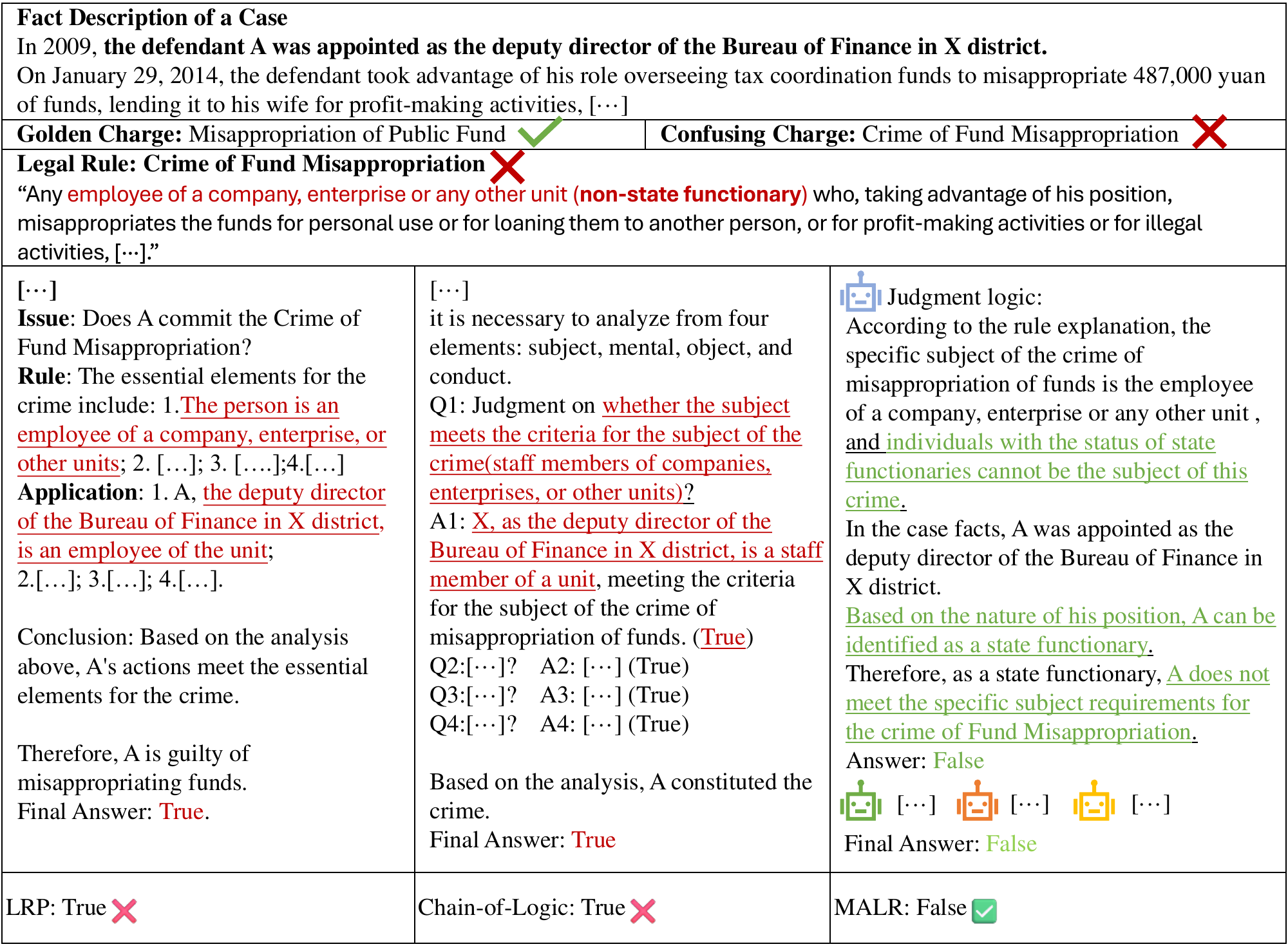} 
  \centering
  \caption {Case study for a given case. The \textcolor{green}{\underline{green}} parts mean are the most critical information for distinguish the confusing charges, the \textcolor{red}{\underline{red}} parts are contents that do not match the facts of the case.}
  \label{fig-casestudy}
\end{figure*}

\noindent \textbf{Ablation Results:}
Table \ref{tab:ablation} demonstrates the effectiveness of the components in adaptive rule-insights training module. (1) The results of ``w/o $E_{success}$ (without Successful Experience)'', ``w/o $E_{esp}$ (without Error-Success-Pair Experience)'', and ``w/o $M_{filtering}$ (without Insight Filtering)'' prove the significance of each designed component in the learning from the trial-and-error process.
(2) The ``\textit{directly generate}'' approach involves encouraging the LLM to generate insights directly based on the legal rules without any training process. However, the performance drops in most situations, sometimes even worse than without using insights at all. A possible explanation is that directly generating insights may lead to the inclusion of unimportant information. We provide case examples with explanations comparing the directly generated insights with those obtained through our training process in Appendix \ref{sec:appendix case insight}.

\noindent\textbf{Open-source LLMs with different model sizes.} 
To further test the applicability of our MALR on different LMs (i.e. different sizes of open-source LLMs), we supplemented relevant experiments using a series of Qwen-2 models~\citep{yang2024qwen2}.
Our findings indicate that our MALR achieves the best results across LLMs of different sizes and adheres to \textbf{scaling laws}.
Interestingly, we also observed \textbf{more significant improvements in smaller LLMs}, which further demonstrates the effectiveness and practical significance of our proposed framework (details can be seen in Appendix \ref{sec:appendix open-source}).

\noindent\textbf{The Challenge and Significance of the Confusing Charge Prediction Task:} To demonstrate the challenge and significance of our proposed task, we thoroughly \textbf{compared General Charge Prediction and Confusing Charge Prediction}. These comparisons clearly indicate that while general legal models perform well on traditional general charge prediction tasks, they are less effective for the confusing charge prediction task.
Additionally, we also \textbf{analyzed the performance of human annotators} on this task. Our findings demonstrate the urgency and importance of this task setting, and they reveal that MALR can even surpass human performance (details can be seen in Appendix \ref{sec:appendix task-challenge}).

\subsection{Case Study}
Figure \ref{fig-casestudy} presents an example of different methods used to predict confusing charges.
As demonstrated in the case, our framework effectively focuses on the most critical aspects of the legal rules and makes a well-reasoned judgment.
In contrast, both LRP and Chain-of-Logic overlook the crucial information in the legal rules, resulting in their failure to accurately predict the confusing charge. 

\section{Conclusion}
In the study, we introduce a challenging task to better evaluate LLMs' capability to comprehend legal theories. The proposed MALR framework can automatically decomposes complex legal tasks and extracts insights from legal rules, enhancing LLMs' legal reasoning abilities. Extensive experiments demonstrate MALR's effectiveness in equipping LLMs with a robust understanding of legal rules.

\section{Ethical Considerations}
The datasets we used for evaluation are all from public legal datasets, and information about the defendants has been anonymized. To ensure personal privacy is not violated, we conduct a secondary review before releasing our dataset to ensure all personal information has been completely removed.

Our work focuses on exploring algorithms to enhance the complex reasoning capabilities of LLMs, rather than replacing human judges or being directly used in real-world decision-making applications. 
In practical use, human judges should act as the final safeguard to maintain fairness and mitigate the potential harms related to algorithms. We will restrict its use for non-commercial purposes such as academic research through a specific license.

\section{Limitations}
Our work has two main limitations.
First, even though we achieved great results, MALR did not predict correctly on all confusing charge pair cases. In the future, retrieval augmented generation could help our model perform better.

Second, our framework shows that LLMs can self-improve by summarize insights into the rules from trials and errors, which helps LLMs to better perform in complex legal reasoning tasks.
Nevertheless, the potential for applying this approach in other fields such as medicine, finance, and scientific discovery remains unexplored. In the future, our framework could be applied in diverse domains.

\section*{Acknowledgments}
This work is supported by Alibaba Innovative Research Program.
This work is supported by the National Natural Science Foundation of China (72104212, 62106039), the Natural Science Foundation of Zhejiang Province (LY22G030002), and the Fundamental Research Funds for the Central Universities.

\bibliography{custom}

\appendix
\section{Prompt Template for our MALR Agents}
\label{sec:appendix MALR Template}
\begin{figure*}[t]
  \includegraphics[width=1\linewidth]{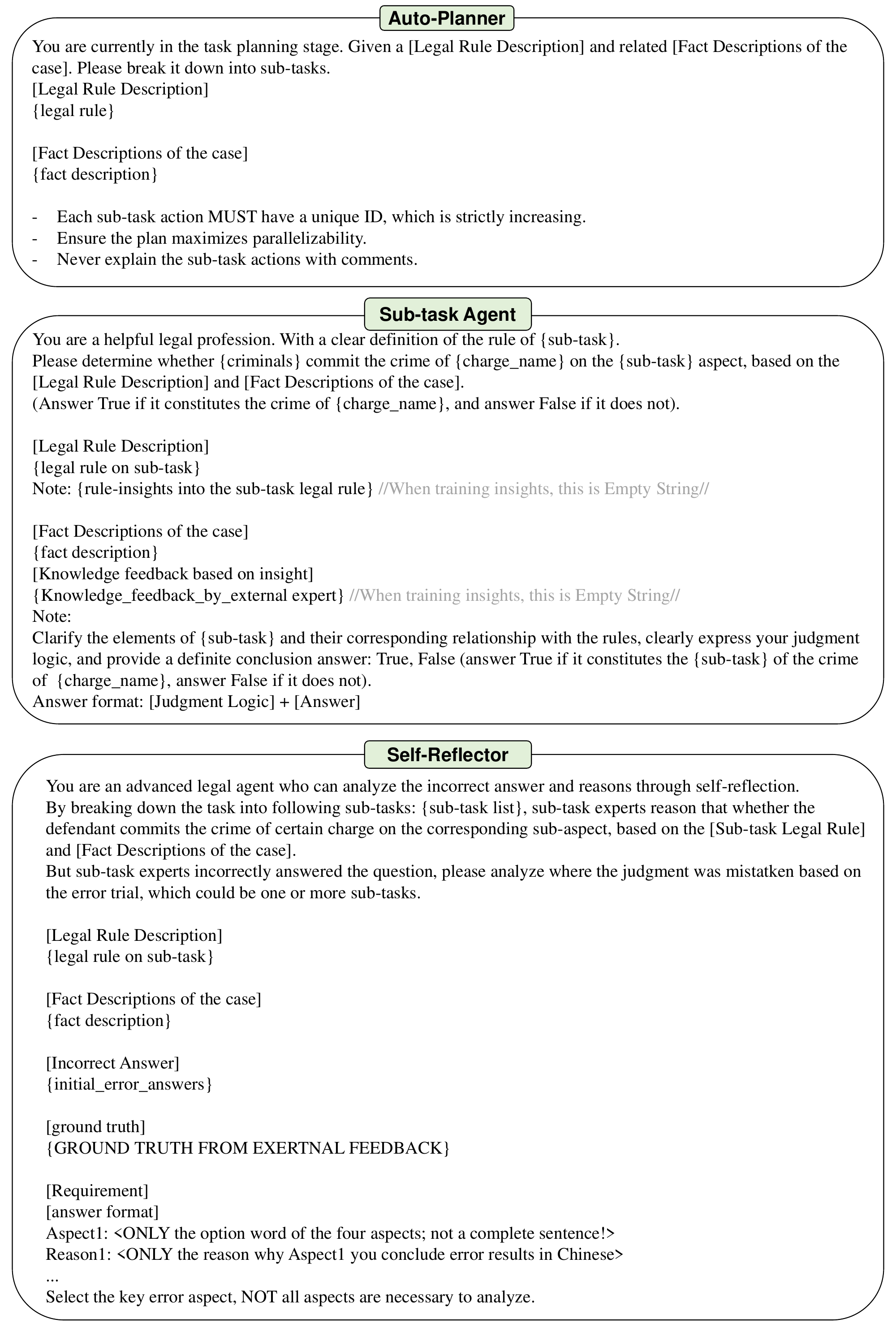} 
  \caption {Prompt Template for Auto-Planer, Sub-task Agent and Self-Reflector}
  \label{fig-malr-prompt-template1}
\end{figure*}

\begin{figure*}[t]
  \includegraphics[width=1\linewidth]{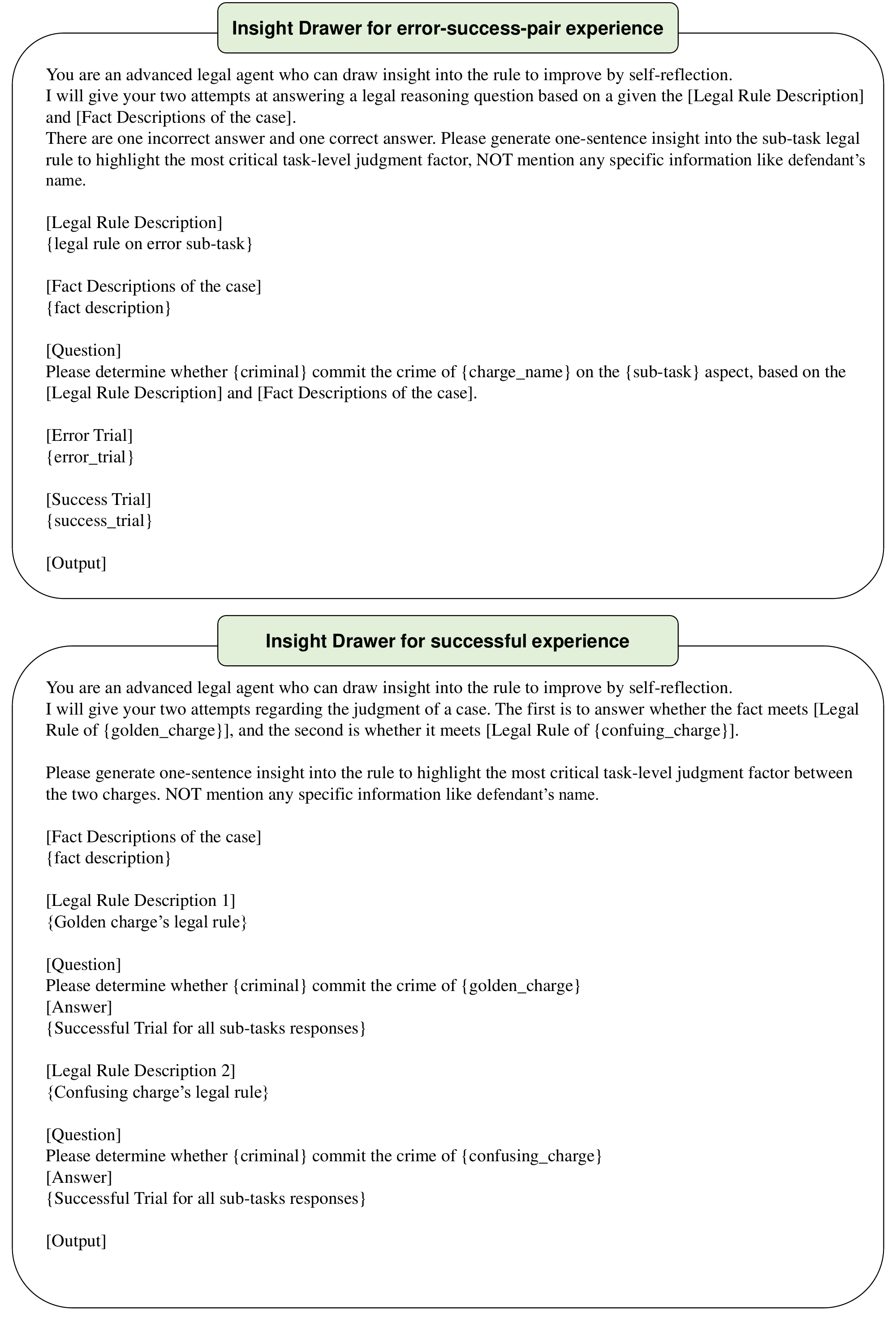} 
  \caption {Prompt Template for Insight Drawer}
  \label{fig-malr-prompt-template2}
\end{figure*}

\begin{figure*}[t]
  \includegraphics[width=1\linewidth]{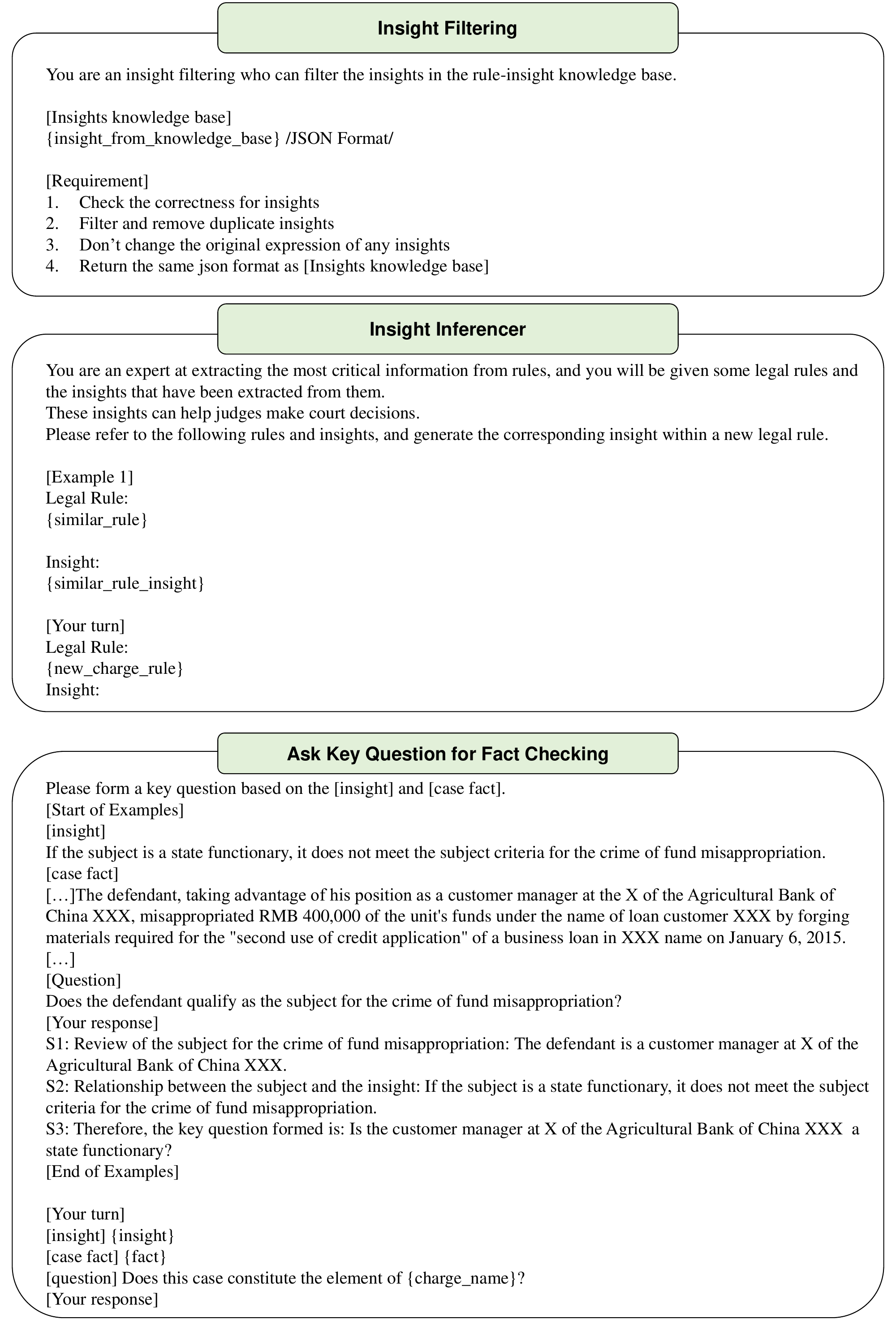} 
  \caption {Prompt Template for Insight Filtering, Insight Inference and Ask Key Question for Fact Checking}
  \label{fig-malr-prompt-template3}
\end{figure*}

The prompt Templates for each agents can refer to Figure \ref{fig-malr-prompt-template1}, \ref{fig-malr-prompt-template2}, \ref{fig-malr-prompt-template3}. We provide prompt templates in English; however, when applied in practice, these templates can be adapted to different languages by translating them into the corresponding language-specific prompts.

\begin{table*}[h]
    \centering
    \scalebox{1}{
    \begin{tabular}{c|cccc}
         \hline
         Methods&  Qwen-2-1.5B&  Qwen-2-7B&  Qwen-2-57B-A14B& Qwen-2-72B\\
         \hline
         ZS-CoT&  \underline{16.3}&  6.25&  16& 21\\
         LRP&  14.8&  6&  13& 26.3\\
         \hline
         FS-Prompt&  7.8&  \underline{29}&  \underline{35}& \underline{48.5}\\
         FS-CoT&  10.8&  12.5&  20.5& 33\\
         Chain-of-Logic&  2.3&  8.3&  15.8& 28\\
         \hline
         MALR w/o insight&  21.8&  21&  22.8& 40\\
         MALR w/o ask&  24&  33.3&  31.5& 48.8\\
         MALR (our)&  \textbf{27.3}&  \textbf{38.3}&  \textbf{40}& \textbf{50.5}\\
         \hline
         Improvement&  67.7\%&  31.9\%&  14.3\%& 4.1\%\\
         \hline
    \end{tabular}
    }
    \caption{Performance of Qwen-2 models with different sizes on CAIL2018 dataset. The "Improvement" shows the performance improvement of our MALR compared to the strongest baseline.}
    \label{tab:open-source}
\end{table*}

\section{Dataset and Experiments Information}
\label{sec:appendix dataset}

\textbf{CAIL2018} is a popular Chinese charge prediction datasets. It consists of real-world cases, each of which includes a fact description and the corresponding charges labels.

\textbf{CJO} is another Chinese legal dataset, same source from the CAIL2018, which is constructed to mitigate the potential data leakage.

\textbf{CAIL-I} contains 462 innocent cases that did not involve any crime. The dataset also has annotations for the criminal charge most similar to the non-criminal facts. The legal judgment prediction for an innocent case adheres to the presumption of innocence. It can evaluate whether LLMs can fully conform to legal rules for reasoning.

\begin{figure*}
    \centering
    \includegraphics[width=1\linewidth]{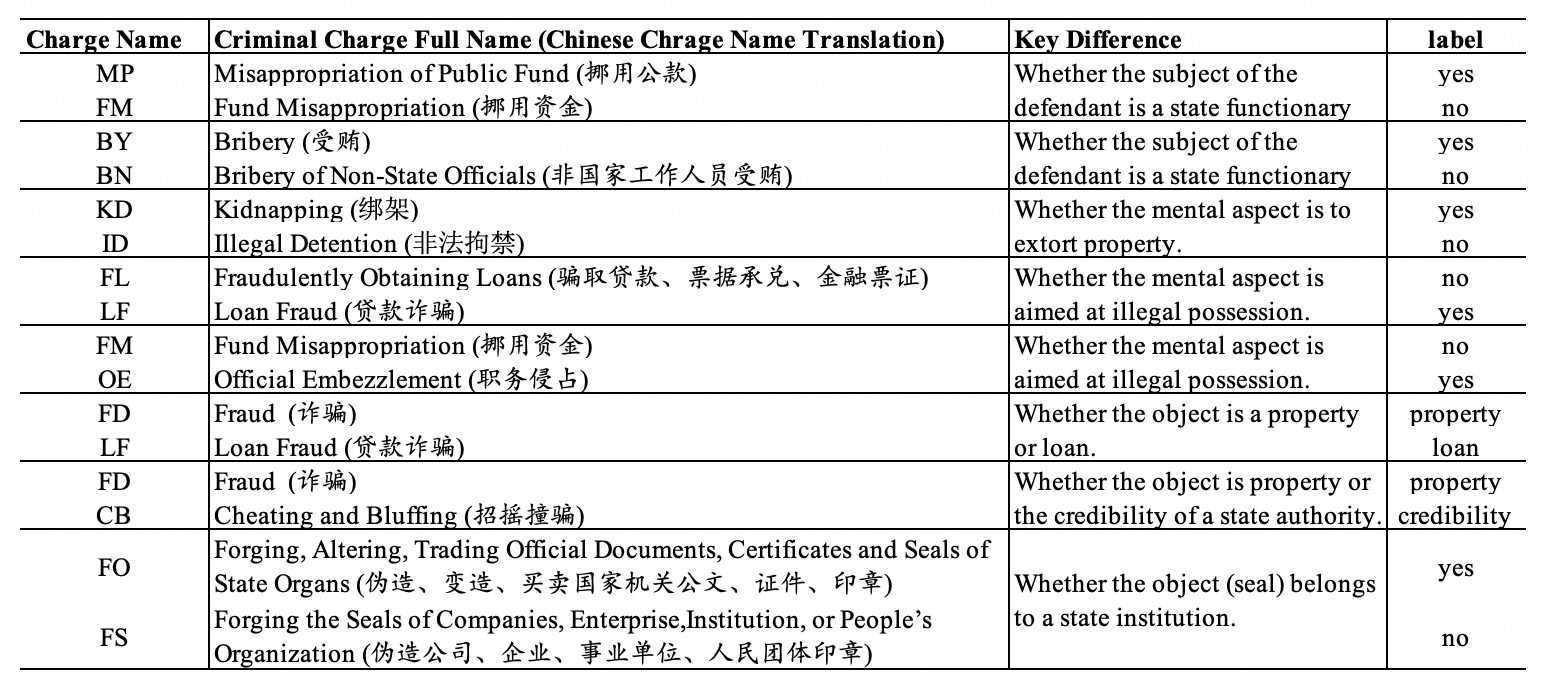}
    \caption{Key difference between each pair of confusing charge}
    \label{fig:confusing chagre}
\end{figure*}

Key differences between each pair of confusing charges are provided in Figure \ref{fig:confusing chagre}.
\textbf{Model Cost}. Statistically, the total token of our method is 1365 for each CAIL2018 example and the inference time per example is about 22s.

\section{Prompt Template for Baseline}
\label{sec:appendix Baseline Template}
\begin{figure*}[t]
  \includegraphics[width=1\linewidth]{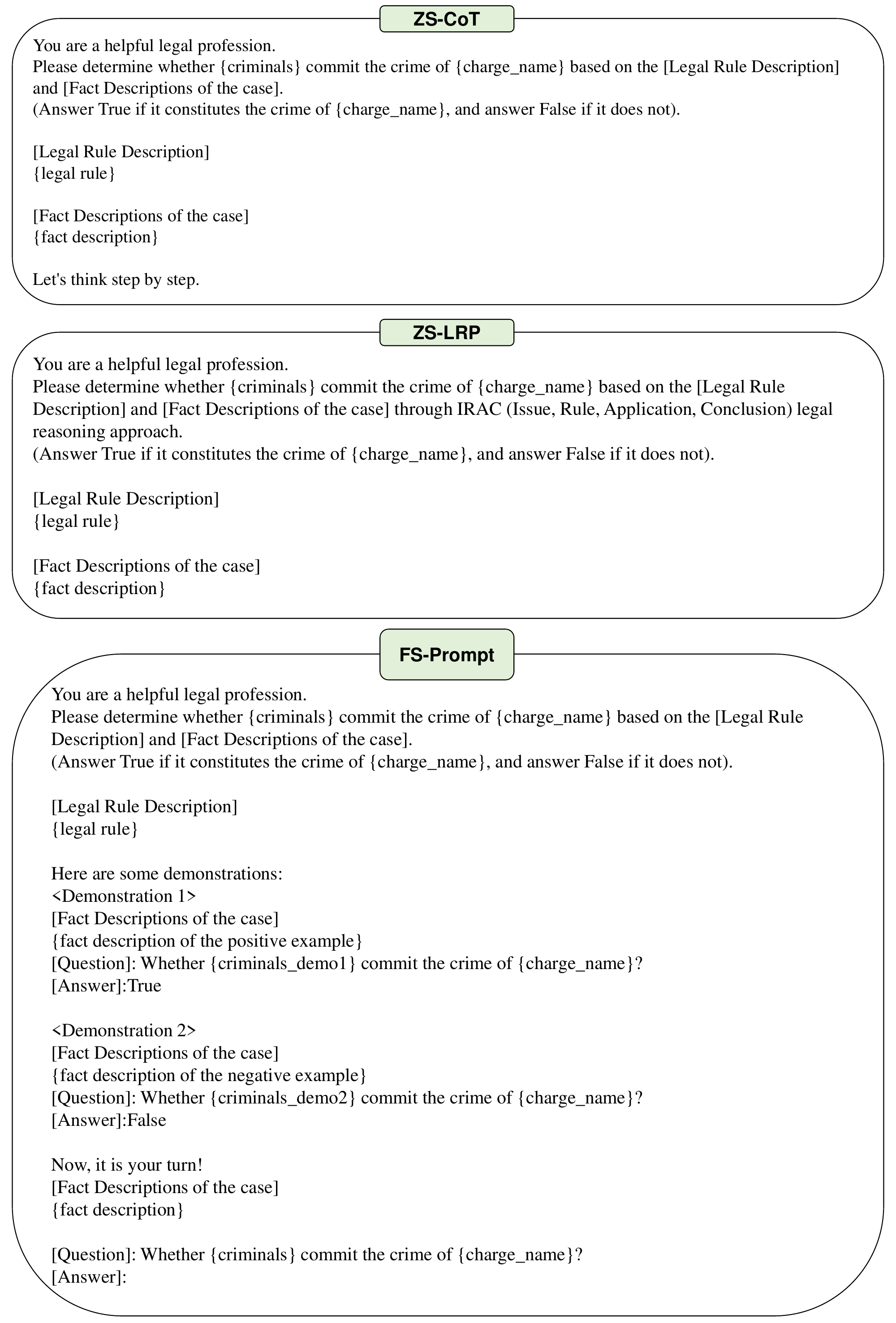} 
  \caption {Prompt Template for baseline ZS-CoT, ZS-LRP and FS-Prompt}
  \label{fig-baseline-prompt-template1}
\end{figure*}

\begin{figure*}[t]
  \includegraphics[width=1\linewidth]{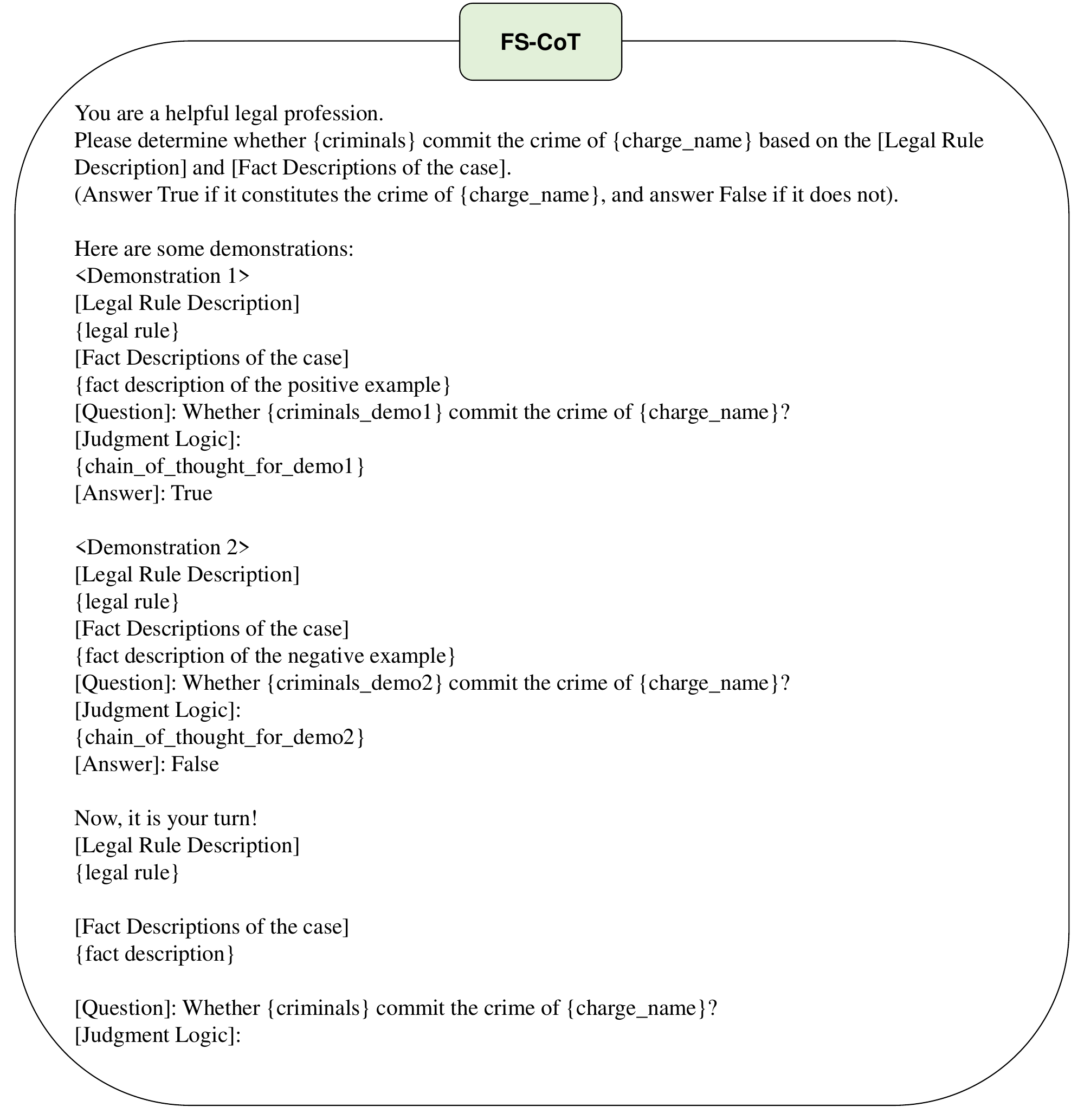} 
  \caption {Prompt Template for baseline FS-CoT}
  \label{fig-baseline-prompt-template2}
\end{figure*}

\begin{figure*}[t]
  \includegraphics[width=1\linewidth]{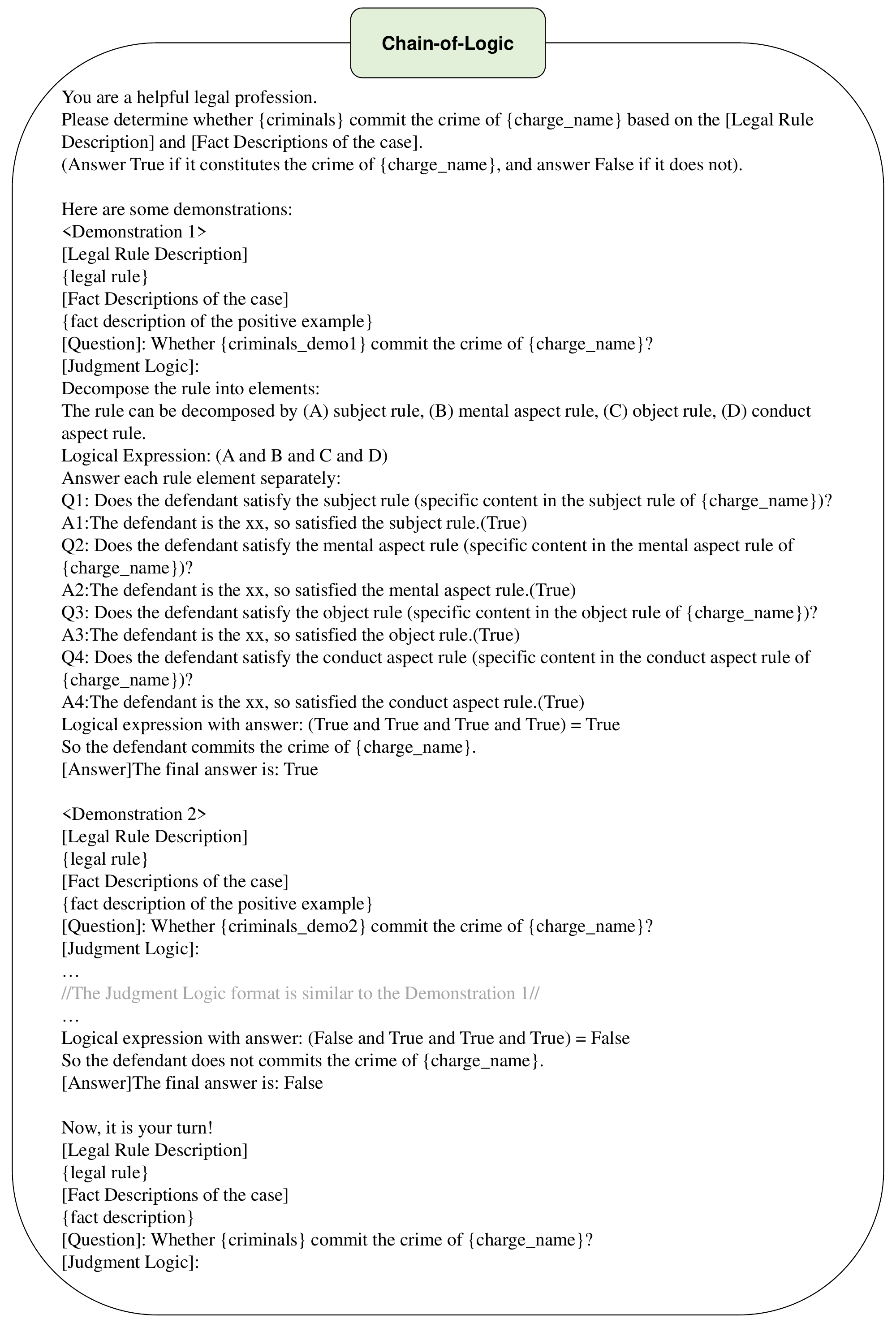} 
  \caption {Prompt Template for baseline Chain-of-Logic}
  \label{fig-baseline-prompt-template3}
\end{figure*}

The prompt templates for each baseline can refer to Figure \ref{fig-baseline-prompt-template1}, \ref{fig-baseline-prompt-template2}, \ref{fig-baseline-prompt-template3}. We provide prompt templates in English; however, when applied in practice, these templates can be adapted to different languages by translating them into the corresponding language-specific prompts.

\section{Specific Performance in the CAIL2018 dataset}
\label{sec:appendix detail}
Table \ref{main-result2} details the specific performance for each confusing-charge pair in the CAIL2018 dataset. The proposed MALR framework achieves the best performance on nearly all confusing charge pairs.

\begin{table*}[p]

  \centering
 
 \scalebox{0.75}{
  \begin{tabular}{c|cc|cc|cc|cc|cc|cc|cc|cc}
    \hline
      Golden Charge& MP& FM& BY& BN& KD& ID& FL&LF&FM& OE& FD&    LF&FD&CB&FO&FS\\
      \hline
\multicolumn{17}{c}{GPT-3.5}\\
\hline
 ZS-CoT& 4.0 & 0.0 & 12.0 & 8.0 & 32.0 & 4.0 & 0.0 & 0.0 & 0.0 & 16.0 & 36.0 & 0.0 & 24.0 & 8.0 & 56.0 &0.0 
\\
      LRP& 0.0 & 0.0 & 0.0 & 0.0 & 32.0 & 0.0 & 4.0 &0.0 &0.0 & 16.0 &                            20.0 &    4.0 &20.0 &4.0 &48.0 &8.0 
\\
      FS-Prompt& 0.0 & 0.0 & 0.0 & 0.0 & 40.0 & 8.0 & 0.0 &0.0 &\textbf{8.0}& 8.0 &                            \textbf{76.0} &    4.0 &\textbf{72.0} &4.0 &\textbf{68.0} &0.0 
\\
      FS-CoT& 8.0 & \textbf{64.0}& 12.0 & 0.0 & 20.0 & 0.0 & 0.0 &0.0 &0.0 & 12.0 &                            36.0 &    0.0 &16.0 &0.0 &24.0 &0.0 
\\
 Chain-of-Logic& 0.0 & 28.0& 0.0 & 8.0& 0.0 & 8.0& 0.0 & 0.0 & 4.0& 0.0 & 4.0& 0.0 & 28.0& 0.0 & 20.0&40.0\\
 MALR (Our)& \textbf{24.0} & \textbf{64.0} & \textbf{64.0} & \textbf{16.0} & \textbf{68.0} & \textbf{28.0} & \textbf{28.0} & \textbf{12.0} & \textbf{8.0} & \textbf{28.0} & 24.0 & \textbf{72.0} & 32.0 & \textbf{44.0} & 52.0 &\textbf{88.0}\\
      \hline
\multicolumn{17}{c}{GPT-4}\\
\hline
 ZS-CoT
& 12.0 & 52.0 & 68.0 & 12.0 & 24.0 & 0.0 & 0.0 & 0.0 & 4.0 & \textbf{40.0}& 96.0 & 4.0 & 96.0 & 8.0 & 76.0 &80.0 
\\
 LRP
& 20.0 & 76.0 & 60.0 & \textbf{32.0} & 16.0 & 44.0 & 8.0 & 0.0 & 28.0 & 24.0 & 80.0 & 8.0 & 80.0 & 16.0 & 56.0 &60.0 
\\
 FS-Prompt
& 12.0 & 56.0 & \textbf{84.0} & \textbf{32.0} & 20.0 & 0.0 & 16.0 & \textbf{60.0}& \textbf{88.0}& 20.0 & 56.0 & 32.0 & 92.0 & \textbf{20.0}& 40.0 &28.0 
\\
 FS-CoT
& 8.0 & 64.0 & 48.0 & 12.0 & 24.0 & 0.0 & 0.0 & 0.0 & 0.0 & 20.0 & \textbf{100.0}& 0.0 & 92.0 & 4.0 & \textbf{84.0}&88.0 
\\
 Chain-of-Logic
& 8.0 & 80.0 & 56.0 & 16.0 & 24.0 & 0.0 & 0.0 & 0.0 & 8.0 & 16.0 & \textbf{100.0}& 0.0 & 92.0 & 12.0 & 80.0 &84.0 \\
 MALR (Our)& \textbf{36.0} & \textbf{88.0} & \textbf{84.0} & \textbf{32.0} & \textbf{36.0} & \textbf{76.0} & \textbf{32.0} & 28.0& 44.0 & 20.0 & 96.0 & \textbf{56.0} & \textbf{100.0}& 12.0 & 72.0 &\textbf{96.0} \\
 \hline
  \end{tabular}

  }
  \caption{\label{main-result2}
    Results on each criminal charge of confusing-charge pairs on CAIL2018 dataset. 
  }
\end{table*}

\section{More Cases for our insights}
\label{sec:appendix case insight}
\begin{figure*}[t]
  \includegraphics[width=1\linewidth]{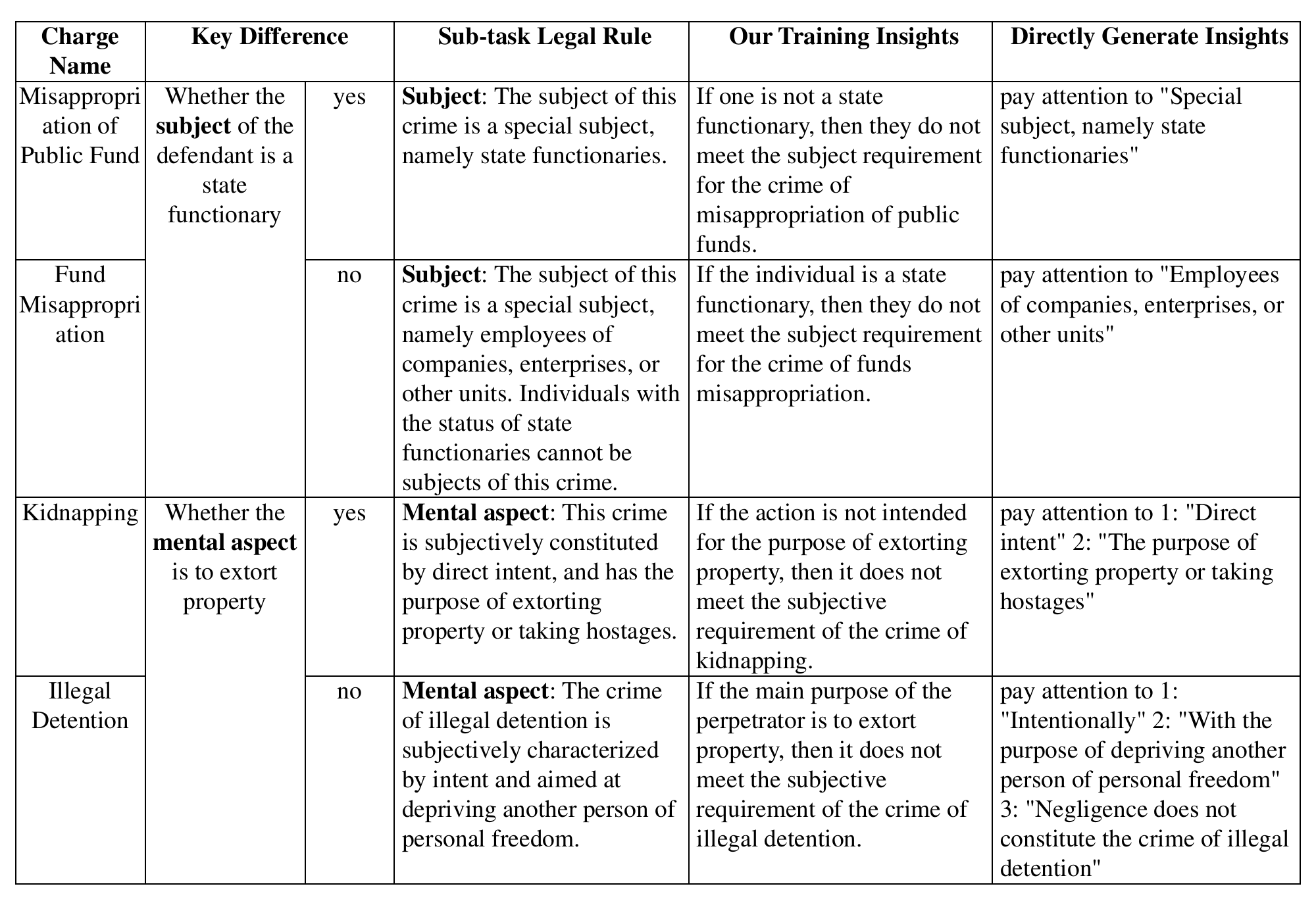} 
  \caption {Case study for illustrating the effectiveness of our training insights.}
  \label{fig-insights-keystudy}
\end{figure*}

Figure \ref{fig-insights-keystudy} shows our training rule-insights can better learn the slight difference in the legal rules, which encourage the LLMs to better understand the legal rules.

\section{MALR performance in Open-source LLMs}
\label{sec:appendix open-source}
We applied Qwen-2~\citep{yang2024qwen2} in our framework for additional test. Qwen-2 includes a series of models with different sizes. We selected models with sizes of Qwen-2-1.5B, Qwen-2-7B, Qwen-2-57B-A14B, and Qwen-2-72B. The results on the CAIL-2018 dataset are as shows in Table \ref{tab:open-source}:

We also observed some interesting phenomena:

\textbf{Scaling Law}: For models of different sizes, performance gradually improves as the model size increases. This is consistent with the conclusions of the scaling law and proves the robustness of our proposed model, which works effectively across models of varying sizes.

\textbf{Significant Improvements for Smaller LLMs}: The MALR method may offer more substantial improvements for LLMs with relatively weaker foundational capabilities. The smallest model, Qwen-2-1.5B, has a 67.7\% improvement over the strongest baseline with our MALR. In contrast, the 72B Qwen model only showed a 4.1\% improvement over the highest baseline. This indicates that our framework can bring significant enhancements when the LLM’s capacity is relatively limited. This is crucial for real-world applications, as not all institutions and individuals can afford the most powerful LLMs.

\section{Challenge and Significance of our proposed task}
\label{sec:appendix task-challenge}
We conducted extensive experiments and human evaluations. The results demonstrate the urgency and importance of our Confusing Charge Prediction task setting and show that MALR can even outperform human performance in our task.

\textbf{Comparisons to General Charge Prediction.}
Confusing Charge Prediction~\citep{CAIL-I} is motivated by the challenge that legal prediction models encounter when distinguishing between charges with similar meanings in real legal scenarios. Compared to General Charge Prediction, Confusing Charge Prediction~\citep{xiao2018cail2018} is significant and valuable because it tackles a critical and practical challenge. In real legal scenarios, legal models often confuse and misinterpret charges with subtle differences. Enhancing the ability of models to accurately predict these confusing charges can improve the accuracy and reliability of legal AI systems.

From the perspective of task formulation, the General Charge Prediction Task is a multi-label classification problem where the input is a fact description and the output is one of the charge categories. Our Confusing Prediction Task can be described as follows: For a given fact description, the prediction model aims to determine if it satisfies the rule of the golden charge while not matching the rule of a confusing charge.

First, based on the typical task setup for General Charge Prediction, we performed a multi-label classification comparison. We trained Lawformer~\citep{xiao2021lawformer} on the entire CAIL-2018 training set. The model achieved an overall accuracy of 85.15\% on the test set. In contrast, the average prediction accuracy for the 8 pairs of confusing charges selected in our paper was only 72.19\% (nearly 13\% lower). For example, for the charge of "Misappropriation of Funds" the model's accuracy was 76.45\%. Among the misclassified samples, 51.35\% were predicted as a confusing charge "Misappropriation of Public Funds". Similarly, for the charge of "Bribery by Non-State Officials", the model's accuracy was 70.54\%. Among the misclassified samples, 58.84\% were predicted as a confusing charge of "Bribery". In contrast, certain charges that are relatively easy to distinguish, such as "Illegal Cultivation of Drug Plants" had an accuracy of 99.95\%, and "Environmental Pollution" had an accuracy of 99.3\%.

Second, We evaluated two small legal language models (legal-xlm-roberta-base~\citep{niklaus2023multilegalpile} and lawformer\citep{xiao2021lawformer}) based on the CAIL-2018 dataset, maintaining consistent confusing charge prediction task settings. The evaluation results were as follows: the accuracy of Lawformer was 2\%, and the accuracy of legal-xlm-roberta-base was 2.25\%.
In contrast, our proposed MALR leverages the generalization and comprehension abilities of LLMs, highlighting the advantages of our framework.

\textbf{Comparisons to Human Annotation.}
To further validate the quality of our proposed task and the effectiveness of the MALR framework, we compared the results of the LLM with those achievable by humans. 
We extracted 20 case facts from the CAIL-2018 dataset and randomly selected either the golden charge or a confusing charge, along with the corresponding legal rules for each case. 
Our human evaluation follows standard LLM assessment practices: ensuring annotators have comprehension and reasoning skills, with minimal prior knowledge of the answers. Therefore, we recruited 6 annotators, all with bachelor’s degrees and no legal background. They will receive the necessary training to complete the tasks, and during the evaluation process, they will be provided with legal rules to aid their reasoning, ensuring consistency with the LLM's input.

The human annotators achieved an average accuracy of 62.5\%, with an average completion time of 28.3 minutes. The LLM baseline methods achieved an average accuracy of 53\%, with an average completion time of 8.2 minutes. The accuracy of our proposed MALR is 65\%, with a total execution time of 10.6 minutes.
Our proposed MALR framework not only surpasses the average performance of human annotators in terms of accuracy but also demonstrates superior efficiency in execution time. These results further prove the effectiveness and efficiency of our methodology.

\end{document}